%% file: Main.tex
\documentclass[10pt,twocolumn,letterpaper]{article}

\usepackage[pagenumbers]{wacv} 

\usepackage{wrapfig} 
\usepackage{cancel} 
\usepackage{graphicx}
\usepackage{amsmath}
\usepackage{amssymb}
\usepackage{booktabs}

\usepackage[pagebackref,breaklinks,colorlinks]{hyperref}

\usepackage[capitalize]{cleveref}
\crefname{section}{Sec.}{Secs.}
\Crefname{section}{Section}{Sections}
\Crefname{table}{Table}{Tables}
\crefname{table}{Tab.}{Tabs.}

\def\wacvPaperID{437} 
\def\confName{WACV}
\def\confYear{2025}

\begin{document}

\title{Quantised Global Autoencoder:\\A Holistic Approach to Representing Visual Data}

\author{Tim Elsner\and Paula Usinger \and Victor Czech \and Gregor Kobsik \and Yanjiang He \and Isaak Lim \and Leif Kobbelt\\
~\\
Visual Computing Institute\\
RWTH Aachen University\\
{\tt\small elsner@cs.rwth-aachen.de / graphics.rwth-aachen.de}
}
\maketitle

\begin{abstract}
Quantised autoencoders usually split images into local patches, each encoded by one token. This representation is redundant because the same number of tokens is spent per region, regardless of the visual information content in that region. Adaptive discretisation schemes like quadtrees allocate tokens for patches with varying sizes, but this just varies the region of influence for a token which nevertheless remains a local descriptor. Modern architectures add an attention mechanism to the autoencoder to infuse some degree of global information into the local tokens. Despite the global context, tokens are still associated with a local image region. In contrast, our method is inspired by spectral decompositions which transform an input signal into a superposition of global frequencies. Taking the data-driven perspective, we learn custom basis functions corresponding to the codebook entries in our VQ-VAE setup. Furthermore, a decoder non-linearly combines these basis functions, going beyond the simple linear superposition of spectral decompositions.\\
We achieve this global description with an efficient transpose operation between features and channels and demonstrate our performance on compression. We further show that our space can help to significantly improve generation.
\end{abstract}

\input{sections/1_introduction}

\input{sections/2_related_work}

\input{sections/3_main}

\input{sections/4_eval}

\input{sections/5_conclusion}
~
{\small
\bibliographystyle{ieee_fullname}
\bibliography{egbib}
}

\clearpage
\setcounter{page}{1}
\setcounter{section}{0}
\renewcommand{\thesection}{\Alph{section}}
\input{sections/6_appendix}

\end{document}

%% file: sections/1_introduction.tex
\begin{figure}
\centering
\includegraphics[width=0.48\textwidth]{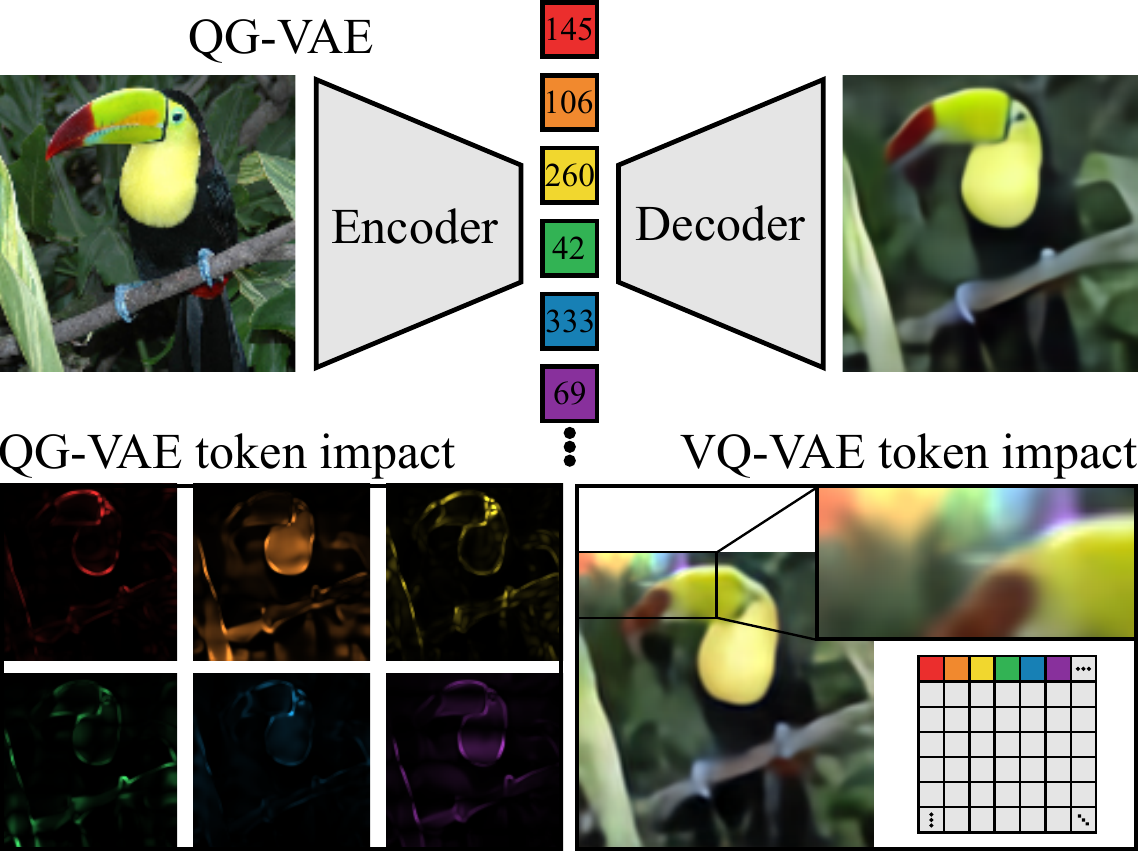}
\caption{Our autoencoder produces a number of \textbf{global tokens} that represent the image, with each token influencing the coloured region of the image. These global tokens help to spread the unevenly distributed information content in the image across the tokens. Example shows $256$ tokens, a ratio of $64$ pixels to one token in $[1,512]$, \ie $288$ byte for an image from ImageNet\cite{imagenet}.}\label{fig:teaser}
\end{figure}
\section{Introduction}
The standard convolution operation on images takes only a local neighbourhood of pixels into account. A vanilla convolutional autoencoder therefore produces an encoding that stores aggregated local information in each latent vector, as enlarging receptive fields becomes costly. Without global information, each part of an image needs the same amount of encodings to describe it properly. Since images often have varying amounts of information content per region, encoding each part of the image with the same amount of tokens is counter intuitive. Some tokens might contain little information, while other regions might not be properly represented because of a lack of tokens in the vicinity.\\

To bypass this problem, recent approaches introduced some degree of global information, often using the attention mechanism\cite{attention} and the following transformers\cite{transformer, vit}. However, even in an attention-infused architecture, latent variables still have a strong local bias. Similarly, even adaptive resolution approaches\cite{hier_vqvae} are improving a low-level grid-based representation, already losing accuracy.\\
In classic image compression, approaches often relied on the Fourier transform to produce an image decomposition where the first few global frequencies provided an approximation of the compressed content. Our approach follows this path, but instead of only providing a linear combination of fixed basis functions to sum up, we learn a function that produces these elements. We provide a strategy to learn a set of tokens with global information that are encoded and decoded together in a holistic fashion, as shown in \cref{fig:teaser}.\\

Our contributions are as follows:
\begin{itemize}
    \item We provide a new architecture recombined from simple parts, but with a small twists to turn local into global information that we then quantise. We do not require any artificial global information aggregators like attention.
    \item We interpret the learned codebooks for the quantisation as Voronoi diagrams that we optimise to evenly distribute information content among tokens.
    \item We turn our approach into a hierarchical decomposition for better interpretability with a simple regularisation.
    \item We demonstrate how our approach can boost performance in downstream tasks like generation, and how it is compatible with autoencoder improvements like sharpening.
    \item We compare our compression rates and provide experiments to better understand our representation.
\end{itemize}
Instead of encoding patches with often vastly different amounts of information content using the same amounts of tokens, our global approach on compression shows better performance, even outperforming methods that learn adaptive hierarchical descriptions\cite{hier_vqvae}. We demonstrate autoregressive generation as a downstream task that greatly benefits from our latent space.\\
We build our approach around the quantisation mechanism introduced by the VQ-VAE\cite{vqvae} framework to obtain quantised tokens. We introduce global information into our tokens with a simple operation: After producing a feature map of the image, we transpose image features with channels, then project each "slice" of the feature map down to a single token.\\
As we introduce global elements into the VQ-VAE framework, we call our approach \textit{Quantised Global Variational Autoencoder (QG-VAE)}.  When adding additional sharpening in the spirit of VQGAN\cite{vqgan}, we refer to the approach as \textit{QGGAN}.

%% file: sections/2_related_work.tex
\section{Related Work}
We discuss \textit{Traditional Compression Methods} for image compression and learned methods based around \textit{Autoencoders}, as we combine ideas from both domains. We further elaborate the different learned autoencoder-based approaches, then dive into \textit{Quantised Methods} that try to increase the information content of the representation, as we aim to produce a quantised and finite representation.
\subsection{Traditional Compression Methods}
In traditional image compression, images are often decomposed into different base frequency bands, then expressed as a linear combination thereof. Exemplary, Watson \etal use the discrete cosine transformation\cite{comp_cosine} that was later used for the JPEG compression standard\cite{jpg}. Similarly, the Laplacian Pyramid\cite{laplacian} expresses an image as a set of frequency images ordered from coarse to fine, finding application \eg in texture synthesis \cite{heeger}. As a related approach, Eigenfaces\cite{eigenfaces} compute a basis set of faces, then express faces as a linear combination, only storing coefficients of the basis functions.\\
These methods are fast and do not require training a neural network, but all only express images as a linear combination of components, never in a context sensitive and holistic manner.
\subsection{Autoencoders}
Neural networks are trained to produce information to output values that fulfil some objective. Reproducing the same output, called Autoencoder, automatically produces a compact representation at the smallest part (bottleneck) of the architecture\cite{first_ae, hinton2006reducing}. Variational autoencoders\cite{vae} extend this concept by introducing a probabilistic take, decoding regions instead of points from latent space to an output.
\paragraph{Quantised Methods}
With the advent of transformers as new state of the art for autoregressive text generation, van den Oord \etal introduced the VQ-VAE\cite{vqvae}, which produces quantised representations of an input by rounding to a nearest codeword from a codebook while minimising the amount of rounding. Similar to VAEs, this maps regions of the latent space that are rounded to the same value, to the same output. This representation allows storing inputs as indices in the codebook, ideal for compression and transformers for generation\cite{dalle,vec_gen,vq_video}. Their quantisation approach forms the basis of our approach.\\
While ideas exist that \eg to reduce the complexity of the approach\cite{simple_vq} or to strengthen the architecture\cite{arch_imp_1, vqvae2}, most related work focuses on either improving information content or improving realism of the output. Tackling the uneven distribution of information similar to us, Huang \etal introduce an adaptive refinement of VQ-VAEs, called DQ-VAE\cite{hier_vqvae}, adding extra tokens for refining details in high information content regions. Other ideas include adding a stochastic element to the tokenisation \cite{svqvae}. A different branch focuses on optimising codebook usage, distributing information evenly across the codebook, \eg directly through optimal transport\cite{vq_optimal_transport}, and strategies like resetting unused codewords\cite{cvqvae} and/or regularising the distribution of codewords\cite{lancucki2020robust, vqwae, wu2020vector}. Our approach follows the codebook reset strategy for simplicity.\\
In a different direction, the VQGAN family introduced by Esser and Rombach \etal \cite{vqgan} and following work\cite{vqgan2, vqgan_better} tries to compensate for the lack of information from the compression by filling gaps with a GAN-like\cite{gan,sgan} network that "sharpens" the output. Recently, TiTok\cite{titok} applies a full transformer architecture based on the Vision Image Transformer\cite{vit} to achieve even more compact codes with good quality.

%% file: sections/3_main.tex
\section{Global Quantised Autoencoder}
\begin{figure*}[ht]
    \centering
    \includegraphics[width=\textwidth]{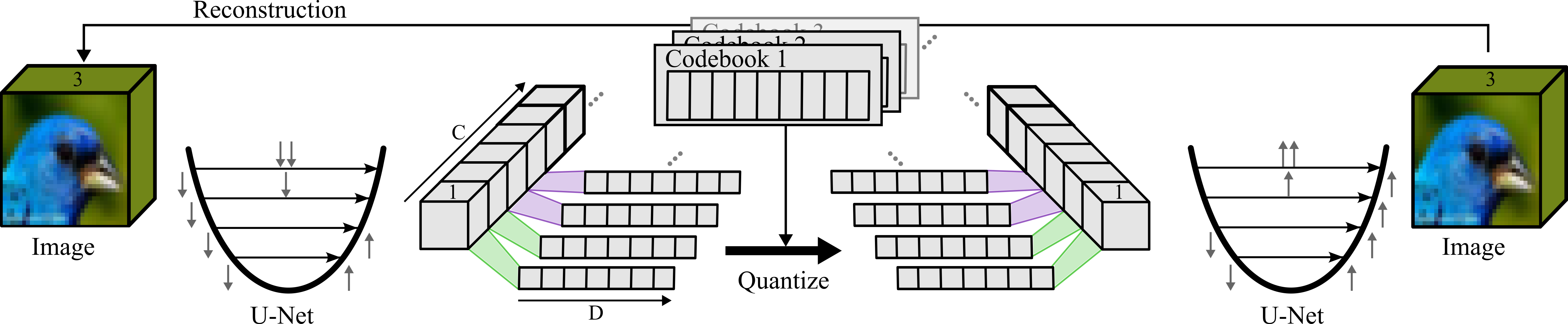}
    \caption{Our base architecture: We encode our data by increasing the input to $C$ different feature maps in the channels (desired number of codewords) with a U-Net \cite{unet}, then compress each channel individually into a token. We do so by transposing channel and collapsed feature dimension, then applying a single linear layer / affine transformation $K$, split into different heads (green and violet). For quantisation, we use the same idea as a naive VQ-VAE. We then decode the result in the same manner.}\label{fig:architecture}
\end{figure*}
\begin{figure}[ht]
\centering
\begin{subfigure}[c]{0.23\textwidth}
    \centering
    \includegraphics[width=\textwidth]{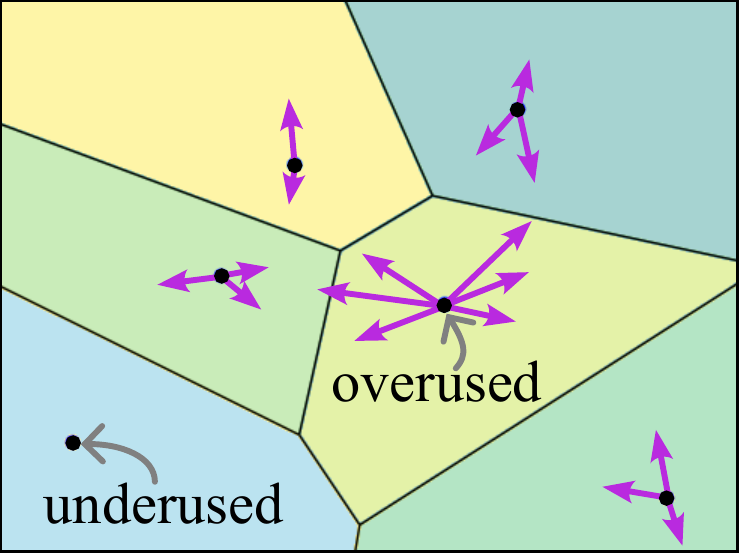}
\end{subfigure}%
\hfill
\begin{subfigure}[c]{0.23\textwidth}
    \centering
    \includegraphics[width=\textwidth]{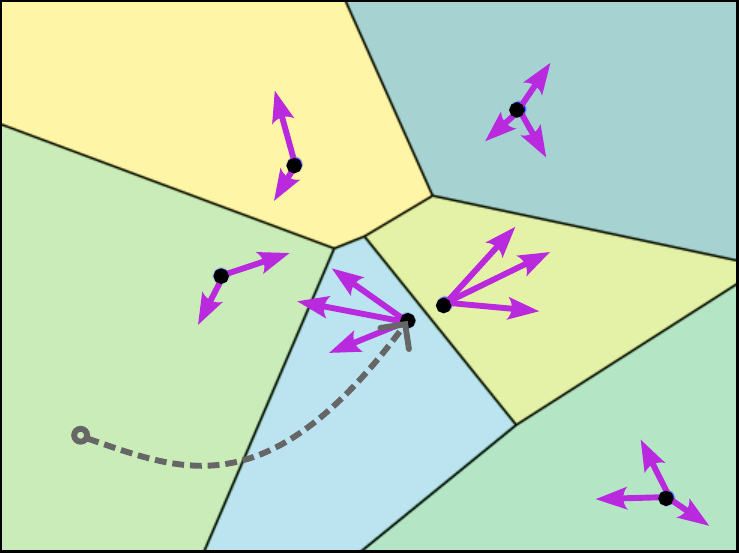}
\end{subfigure}%
\caption{We express re-initialising codewords while relieving overused codewords as re-arranging a Voronoi diagram through moving the Voronoi centres.}\label{fig:reassignment}
\end{figure}
We propose to decompose an input into global tokens unlike a regular VQ-VAE \cite{vqvae}, where each token (mostly) corresponds to a local patch. Exemplary for images, regions with lower amounts of information content do not waste tokens, \eg for an image of a person in front of a blue sky, regions with a blue sky have less information than the face of the person, hence should not use the same amount of tokens. Our approach works in five steps to tackle this problem, the full pipeline shown in \cref{fig:architecture}: 
\begin{enumerate}
    \item The input is processed into a multi-channel feature map of the image, each channel holding global information. 
    \item Each individual channel of the feature map is processed into a single vector that holds global information.
    \item Each vector is quantised into a discrete token.
    \item Every token is decoded back into a single channel of a feature map.
    \item The feature map is processed back to a full image.
\end{enumerate}
Some examples are shown in \cref{fig:ae_examples}, with more shown in \cref{app:extra_results}.
\subsection{Background}
To better motivate the setup of our autoencoder architecture, we first recap some basics about
the Fourier transform.\\
The idea of this transform is to represent a (spatial) signal as a superposition of frequency
functions. If we restrict the signal to be discrete (and periodic), the frequency spectrum is also
discrete (and periodic) leading to the so-called \textit{Discrete Fourier Transform (DFT)}:
\begin{equation}
F(u, v) = \sum_{x=0}^{W-1} \sum_{y=0}^{H-1} I_{x,y} \cdot e^{-i2\pi \left(\frac{ux}{W} + \frac{vy}{H}\right)},
\end{equation}
where $I_{x,y}$ describes a pixel of an image $I$ of the size $W \times H$. By using (complex) sine and cosine functions as basis functions of a linear function space, each input signal can be effectively approximated by a linear combination of frequency basis functions. The coefficients of this linear combination are the frequency components (complex amplitudes). If we reconstruct the signal from only a sub sequence of the frequency components, we obtain a low frequency approximation of the input.
More generally, we can express the frequency decomposition $F$ for $n$ frequencies as a set of independent frequency band computations with:
\begin{equation}\label{eq:general_decomposition}
F = \{f(I, i) | i=0,..,n\}
\end{equation}
These can be decoded back to an image by linear combination of an inverse function $f^{-1}$:
\begin{equation}\label{eq:general_decomposition_decode}
    \hat{I} = \sum_{i=0,..,n} f^{-1}(f(I, i), i)
\end{equation}
While this is a generally good representation of arbitrary data, we rather want to learn a frequency decomposition-like representation that is \textbf{content sensitive} and \textbf{more expressive}. We do so by learning individual functions per frequency, \ie not only by modulating the frequency of a sine/cosine function, but by training a neural network that applies an individual function per frequency band. We further want to allow more complex relationships between the frequency bands, \ie not only sum up the frequencies as a linear combination, but to process them in context to each other. With both the global tokens and the non-trivial combination of frequency bands for decoding, we describe our approach as \textit{A Holistic Approach to Representing Visual Data} and call it \textit{Quantised Global Variational Autoencoder (QG-VAE)}. 
Hence, \cref{eq:general_decomposition} and \cref{eq:general_decomposition_decode} become:
\begin{equation}
    F = E(I), \hat{I} = D(F) = D(E(I))
\end{equation}
for a learnable function $E$ and its inverse $D$ being holistic functions in the sense that they process all frequency bands together.\\
We call the resulting factors of our decomposition \textit{pseudo frequency tokens}, as they differ to a regular frequency decomposition in two main aspects. While in the Fourier transform, the coefficients are computed by evaluating a dot product between the input and a fixed frequency basis functions, we replace this by a learned neural network and quantise these coefficients. While learning an individual function per pseudo frequency would be ideal, this would, for $C$ frequency bands, require learning $C$ different global functions that process the input image. This is impractical because a high number of parameters and hence difficulties in generalising across multiple images. It also does not allow processing the different pseudo frequencies in context to each other. Instead, we resort to a small change that makes our approach feasible. Where DFT applies a different function to the same image through the changed exponent to produce each frequency, we instead use the same (learned) function $K$ at our core to produce the different pseudo frequencies. We obtain these pseudo frequencies $S_i, i \in {0, 1, ..., C}$ by applying $K$ to $C$ different feature maps of the input image, with $E$ producing a feature map of the input:
\begin{equation} 
    E(I) = F_{map}, S_i = \{ K(F_{map}^i)~|~i~\in {0, 1, ..., C}\}
\end{equation}
With $F_{map}^i$ describing a single channel of the feature map.
In our case, $K$ are always learned affine transformations turning $W \cdot H$ dimensional feature maps into $64$-dimensional tokens to quantise, with $K^{-1}$ then transforming $64$-dimensional tokens back into a $W \times H$ dimensional feature map. This feature map is then transformed back into an image by the decoder $D$.\\
Hence, while DFT applies a function with different parameters to the same global image multiple times, we split this in two parts to save learnable parameters. We first apply our learnable transformation $E$ to produce \textit{different} feature maps (instead of the same input for Fourier) that we then put into the same learnable function (instead of applying a individually modulated function for Fourier). This effectively splits the learnable parameters in an individual part and a shared part, making our approach easier to train.\\
We further discuss alternatives to our formulation and their pitfalls in \cref{app:alternatives} (Appendix).
\subsection{Quantised Global Autoencoder}
Our approach follows the same principle for all modalities and variants. We first apply a U-Net \cite{unet} to increase the number of channels of our image to the number of codewords $C$ we want to obtain (\textit{Encoding}), \eg for RGB images processing $[B \times 3 \times W \times H]$ into $[B \times C \times W \times H]$. For closeness to actual code and optimisation, we use a notation with batch size $B$.
Each of these $C$ feature maps (channels) then contains a full image, with each pixel being enriched with global information. We then \textit{switch from local to global features}, compressing each feature map individually into a token, obtaining $C$ tokens.\\
We reverse this compression back into an image with $C$ channels and apply a second U-Net to re-combine these images into one output (\textit{Decoding}).\\
We now describe these steps in more detail.

\paragraph{Encoding and Decoding of the Data}
To ensure we distribute all the information across the feature maps that each form their own pseudo frequency, we apply a U-Net \cite{unet} before and after the switch from local to global features, referred to as Encoder $E$ and Decoder $D$. These U-Nets, together with the affine transformation $K$, learn what corresponds to the projection to the basis functions. To keep memory requirements reasonable, we omit some of the upscaling (encoder) or downscaling (decoder) steps for larger images, \eg turning our $[B \times 3 \times W \times H]$ dimensional input into a $[B \times C \times \frac{W}{2^{f}} \times \frac{H}{2^{f}}]$ feature map for a downscaling factor $f$.

\paragraph{Switching from  Local to Global Features}
To process the multiple feature maps in parallel, we use a transpose operation that swaps feature dimension of the combined width and height with the channel dimension, see \cref{fig:architecture}. For a processed image of size $[B \times C \times W \times H]$, we collapse the two spatial dimensions into one feature dimension, then transpose features and channels to obtain a $[B \times W \cdot H \times C]$ dimensional tensor. Now each of the $C$ features holds a full (flattened) feature map in its channels. We then apply a $1$D convolution with kernel-size $1$, effectively a single learned affine transformation. This reduces our image-sized channels down to the space in which we apply quantisation, typically $64$ dimensions. For this example, we obtain a $[B \times 64 \times C]$ dimensional tensor.  Inspired by transformers, we optionally split the affine transformation $K$ compressing the feature maps into the quantisation space into multiple \textit{heads}, by using multiple different linear layers, \eg one to produce codewords $1$ to $32$, another to produce codeword $32$ to $64$, and so on.\\
To quantise, we use the same approach as traditional VQ-VAEs \cite{vqvae}: Rounding to the nearest codeword from our codebook, and then both regularising so that the encoder produces an output close to the codewords ($L_{comm}$) and so that the codewords move closer to what the encoder produces ($L_{code}$). We use the same loss formulation that uses the distance function to nearest codeword $d$ from the codebook $e$:

\begin{multline}
L = L_{rec} + L_{comm} + 0.25 \cdot L_{code}\\
= [ \hat{D} ( \hat{E} (I) ) - I]^2 + d( \lfloor \hat{E}(I) \rfloor, e) + d( \hat{E}(I), \lfloor e \rfloor)
\end{multline}
where $\hat{E}$ and $\hat{D}$ already contain the affine transformations into quantisation space $K$ and $K'$ and the encoder/decoder $E$ and $D$.

\subsection{Optimising Codebook Useage}
Similar to approaches like SVQ-VAE\cite{svqvae} and VQ-WAE \cite{vqwae}, we want to improve the codebook usage. The VQ-VAE losses do not penalise unused codewords, only producing encoder outputs distant from the codebook. Hence, the autoencoder wastes potential. While one single codebook already provides good results, we use \textit{individual codebooks per token} for better codebook usage and performance with slightly longer convergence time. We use an individual codebook to quantise each feature map $C$, because as opposed to a traditional VQ-VAE, we no longer have the same meaning for tokens across the quantised domain: Multiple similar patches of grass in a quantised image would traditionally be mapped to the same token. However, in our case, the $ith$ codebook index should have a different meaning in the $nth$ frequency band than in the $mth$ frequency band. Note that this does not affect the tokens we need to store, as we still only need to store the list of indices of our tokens to store an input.\\
While resetting unused codebook entries or regularising the distribution of codewords \cite{lancucki2020robust, wu2020vector, cvqvae} is nothing new, the individual nature of our entries (each frequency band having its own codebook) is making this more complicated: 
With individual codebooks per pseudo frequency, we only have few samples for every codebook per batch, making regularisation unstable. We instead opt to reset \textit{underused} codewords to explicitly disencumber \textit{overused} codewords. For this, we consider the space to quantise as the space of a Voronoi diagram, with different codewords being the Voronoi centres.
We can compute codebook resets by simply tracking the gradient magnitude of our codebook over multiple batches as the gradient is a direct measure for how much an entry should be changed to improve the result. Through a number of steps, usually $500$, we sum up all the gradient magnitudes for each codeword. In result, this basically gives each Voronoi cell an error rate. Unused codewords and their associated Voronoi cells will have an accumulated gradient magnitude of $0$, while overused codewords that are often used but are not chosen perfectly will have a high amount of gradient magnitude. Hence, we move an unused codeword (and hence, the associated Voronoi centre) to the position of the most overused codeword with the largest gradient, and perturb it in a random direction $d \cdot \epsilon$. Codebook and commitment loss will then separate the two codewords further, splitting the overused codeword / one overused Voronoi cell into two distinct ones. This balances the amount of information content between the codewords and thus the information content in each Voronoi cell of the encoding space with just a few lines of code.
\subsection{Obtaining a Meaningful Decomposition of the Latent Space}\label{sec:dropout}
While our approach already provides a decomposition into pseudo frequencies that is somewhat structured,  we can order this from important to unimportant. This extra structure can be an additional help to understand our representation and is yet again inspired by classic frequency decomposition.\\
In fields like texture synthesis, a coarse to fine structure for a frequency decomposition is often achieved through the Laplacian pyramid $L$ with levels $0$ to $n$, build from the Gaussian Pyramid $G$ and its levels $G_i$ for an input $G_0 = I$: 
\begin{multline}
G_{i+1} = \text{reduce}(G_i) \quad \text{for} \quad i = 0, 1, \ldots, n-1\\
L_i = G_i - \text{expand}(G_{i+1}) \quad \text{for} \quad i = 0, 1, \ldots, n-1\\
L_n = G_n
\end{multline}

The produced image pyramid thus contains a base image $G_n$ and multiple frequency bands $L_{i-1}$ to $L_0$ that always ensure full reconstruction. When applying the Laplacian pyramid for all image sizes, we are left with $i$ frequency bands. With this formulation, taking the first $k$ frequencies is always the optimal solution in the sense that there is no other way of combining $k$ different of these frequency bands to obtain a better $L2$ error.\\
Inspired by this, we re-formulate our reconstruction loss such that using the first $k$ pseudo frequency bands should provide the best $L2$ loss:\\
\begin{equation}
L_{rec} = \sum_{k \in [0, C]}~(f_k(x) - x)^2
\end{equation}
where $f_k$ is our trainable network only using the first $k$ pseudo frequencies.\\
As training to properly decode this optimally for every input with every different pseudo frequency is cumbersome, we instead do this in a probabilistic way. We apply a dropout-like procedure on every third batch, multiplying everything from the $kth$ latent code onwards with $0$, where $k \sim \{0, 1, 2, \ldots, n\}$:
\begin{equation}
L_{rec} = (D(E(x) \circ d^k) - x)^2
\end{equation}
where $d^k$ is a tensor where $d^k_i = 1$ for $i < k$ and $d^k_i = 0$ otherwise, nullifying every token with an index greater than $k$ of our encoder output. With optimal training, this is close to the formulation of the Laplacian pyramid: Our decoder outputs an average over the dataset when decoding a latent space with no information ($k = 0$), with each pseudo frequency band adding the largest amount of detail possible.
We visualise the resulting decomposition in \cref{fig:dropout}.\\
With this regularisation, we gain more interpretability, but do lose some precision. Instead of enforcing a certain way for our U-Net to work (here expressed as to being able to properly decode with only subsets of our pseudo frequency bands given), we achieve the best performance by processing all pseudo frequencies in a holistic sense.

%% file: sections/4_eval.tex
\input{sections/figures/examples}
\input{sections/tables/compression}
\input{sections/figures/qggan}
\section{Evaluation}
We run all our benchmarks on a single Nvidia GeForce RTX 2080 Ti, demonstrating that our approach requires no expensive hardware or excessive amounts of VRAM. We further use a U-Net with residual blocks from \cite{imagen} and downscaling/upsampling instead of max pooling/transposed convolution to make our code less intricate. We provide our code as simple, one-clickable Jupyter Notebooks on GitHub\footnote{https://github.com/DaiDaiLoh/QG-VAE}.
\subsection{Benchmarks in Compression}
To evaluate the compression ability at the core of our approach, we expand the comparison of VQ-WAE \cite{vqwae}. We train our approach for the same number of epochs ($70$ for CelebA, $100$ for every other benchmark), then evaluate on the test set. Details about our training can be found in \cref{app:benchmark_params}. Results can be found in \cref{tab:compression}, showing the strong performance of our approach. We further like to point out that our approach surpasses most of these benchmarks and saturates earlier.\\
We interpret these results in the context of global versus local descriptions: An image is no longer composed of patches that possibly have visible seams that can be picked up by the NN-based metrics like FID and LPIPS. Our approach is biased towards learning an accurate, holistic representation through its global feature maps instead of local patches that form a token. We also observe that features in produced images compared to a regular VQ-VAE might still not be correct, \ie offset or wrong, but they look much sharper than at similar PSNR values. This heavily shows in the FID scores of CIFAR-10. For reference, consider high-frequency details like the legs of the horse rider in \cref{fig:ae_examples}. Also note that to compute the FID values, the latent space of a model trained on ImageNet is used, which does not contain close up features of faces, and should hence be taken with a grain of salt on the CelebA benchmark. We provide extra qualitative results in \cref{app:extra_results}, in particular on larger datasets, resolutions, and number of tokens. We also do some qualitative analysis and compare to a VQ-VAE directly to better show the visual improvements of our QG-VAE.
\begin{table}[h]
\centering
\begin{tabular}{|c||c|c||c|c|}
\hline
~ &
\multicolumn{2}{|c||}{\textbf{CIFAR-10}\cite{cifar}} &
\multicolumn{2}{|c|}{\textbf{CelebA64}\cite{celeba}} \\
\hline
Metric & DQ & QG & DQ & QG \\
\hline
SSIM $\uparrow$ & 0.499 & \textbf{0.683} & 0.836  & \textbf{0.844}\\
PSNR $\uparrow$ & 17.35 & \textbf{21.94} & 25.48 & \textbf{26.74}\\
LPIPS $\downarrow$ & 0.469 & \textbf{0.357} & 0.190  & \textbf{0.184}\\
FID $\downarrow$ & 140.3 & \textbf{88.35} & 40.74 & \textbf{34.66}\\
\hline
\end{tabular}
\caption{Comparison of training a hierarchical DQ-VAE\cite{hier_vqvae} compared to ours for $50$ epochs, always with our QG-VAE using the same number of tokens ($20$ for CIFAR, $80$ for CelebA64).}\label{tab:yanjiang}
\end{table}
\subsubsection{Comparison to Adaptive Refinement}
We further compare to the approach of Huang \etal \cite{hier_vqvae}. They propose a hierarchical approach to VQ-VAEs by additionally refining those regions that have a high reconstruction error. We applied their code to CIFAR and CelebA, then compare this explicit hierarchical decomposition against our approach in \cref{tab:yanjiang}. As we are concerned for the actual amount of information encoded, not for the qualities of an additional post processing, \eg through a GAN\cite{vqgan}, we chose to use both their approach and our approach \textit{without} this sharpening. We report better performance, concurring with what we also observed in other experiments: With fewer tokens, our approach outperforms other approaches even stronger. While their approach improves results, it can only make a binary decision to refine an area, and due to the locality of the additional tokens, is still bound to a (refined) grid layout.
\subsection{Useability of our Latent Space}
To show that our autoencoder framework is both compatible to common extensions and useful for downstream tasks, we demonstrate \textit{sharpening} and \textit{autoregressive generation}.
\paragraph{Sharpening}
For image compression, high-frequency details are often lost in compression, \eg the scales of a fish are not reconstructed properly. While reconstructing these high-frequency details is not feasible, approaches like VQGAN\cite{vqgan} produce new high frequency details by adding an additional discriminator to sharpen the output, possibly inventing similar details like the ones lost in compression. To demonstrate that our approach is just as flexible as a regular autoencoder when it comes to compatibility with other approaches, we train a VQGAN\cite{vqgan}-inspired approach that uses our autoencoder as its backbone, which we further refer to as \textit{QGGAN}. Examples can be found in \cref{fig:images_qggan} and \cref{fig:extra_images_imagenet}. We report an FID of $11.42$ for our QGGAN and $57.16$ for our trained VQGAN on ImageNet128 with only $256$ tokens after training for $15$ epochs, both without attention. We attribute the difference to more global features, that not only increase compression, but also make the recognition of global context easier to refine small details (\eg refining the surface of a cat as fur once the network recognises the cat).
\begin{figure}
    \centering
    \includegraphics[width=0.48\textwidth]{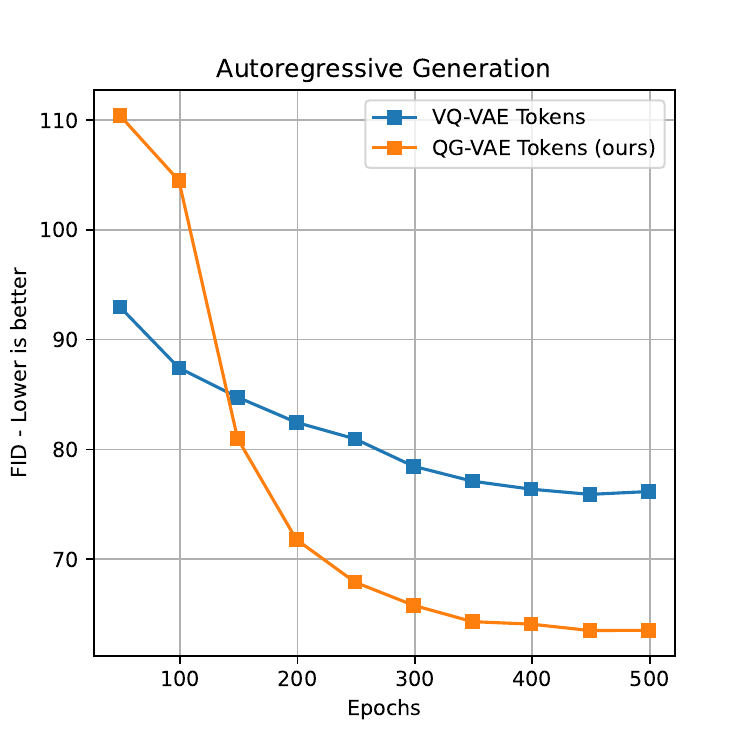}
    \caption{Generation quality when training the same transformer on our pseudo frequency tokens of our QG-VAE on CIFAR-10\cite{cifar} versus training on tokens produced by a VQ-VAE\cite{vqvae}. Our approach performs better, producing lower FID. We compute on $10,000$ unconditional samples each.}\label{fig:app_generative_plot}
\end{figure}
\paragraph{Generation}~\\
\raisebox{0.2cm}{\raisebox{-0.2cm}{\includegraphics[width=0.48\textwidth]{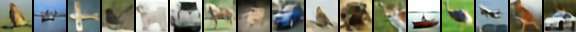}}}
While compression is our main focus, we did some basic exploration of using our tokenised $1$D latent space for generation (examples in the inset). We measure that our natively linear latent code is beneficial for generation in \cref{fig:app_generative_plot}: Autoregressively predicting token by token, and hence, patch by patch of in a $2$D grid, does not come native for a model designed for 1D sequences. We provide results on generating CIFAR images with an autoregressive transformer\cite{mingpt}, hinting at the potential for future work. After an initial delay, which we attribute to more complex relations in global instead of local tokens, \eg predicting a blue sky token next to a blue sky token is initially trivial, the same transformer on tokens of our QG-VAE approach vastly outperforms the VQ-VAE tokens. We show additional examples and explanations in \cref{app:generation}.
\input{sections/figures/dropout}
\input{sections/figures/featuremaps}
\subsection{Interpretation of our Holistic Representation} We generally believe our better compression rate to be rooted in holistic instead of local descriptions of an input. With additional experiments, we demonstrate how our learned latent space produces global features instead of localised ones with additional experiments, from which we can draw further conclusions about the properties of our learned space.\\
\textbf{First}, we visualise the feature maps, \ie the $C$ many different channels of our $[B \times C \times \frac{W}{2^f} \times \frac{H}{2^f}]$ tensor before compressing each $C$ dimension into a token by affine transformation and quantisation in \cref{fig:feature_maps}. This shows that each token captures different global properties, following the holistic strategy to not just encode an input through local patches.\\
\textbf{Second}, we average the change for all possible replacements for a token in the output in \cref{fig:teaser}. We observe very global changes for single token changes, compared to very local changes for VQ-VAE. Similarly, we can slightly change an input image to cause large changes in the learned latent space: For a simple change of $20$-by-$20$ pixels in an $128$-by-$128$ pixel image, we observe a change of $27$ tokens, while we have $\frac{128^2}{256} = 64$ pixels per token. An example can be found in \cref{app:smaller}.\\
\textbf{Third}, we train an ordered latent space according to \cref{sec:dropout}, with the first $k$ tokens always holding the largest amount of content in the least squares sense. Then, we show the learned decomposition in \cref{fig:dropout}. This demonstrates how gradually the image changes from an average to the target image.\\
Notably, even with this enforced decomposition, our approach has properties that are not typical for a simple linear combination. As an example, we consider \eg Eigenfaces\cite{eigenfaces} that compute an ordered set of face coefficients from high to low importance via PCA. Hence, adding new coefficients of base elements provides additional detail, while our approach is able to completely change attributes instead (\eg gradually moving the smile with more tokens in \cref{fig:dropout}). However, our ordering regularisation means that each first $k$ tokens of pseudo frequency bands must also be interpretable on their own. From this, we observe some decay in reconstruction quality compared to our holistic version without regularisation.\\
\textbf{Fourth}, we present a comparison of different stages of the training process for a VQ-VAE and our approach in \cref{app:training} (Appendix).\\
\textbf{Lastly}, we show a small ablation study to demonstrate the impact of the different parameters \cref{fig:ablation}, always showing deviation from the settings we found to be ideal.
\subsection{Limitations, Discussion, Future Work}
While our approach has a number of advantages despite its simple nature and lack of more complex components like attention, we also highlight a number of smaller drawbacks: As tokens are no longer local, changes to a small region will possibly change a large number of tokens, and changes to a single token can impact whole regions. This makes \eg inpainting-like operations more difficult.\\
Further, QG-VAE does generally require more parameters, as two U-Nets as backbone are more expensive as \eg convolution and pooling of a regular VQ-VAE. However, especially for smaller resolutions (\eg CIFAR), we find that our approach performs similarly with the same number of parameters. Even with those higher numbers of parameters, we find that a U-Net with at most $128$ channels is not extra ordinarily expensive given the shown performance, in particular considering that our vanilla loss function identical to a regular VQ-VAE\cite{vqvae} is much cheaper to compute than \eg the mathematical regularisation of \cite{vqwae}. We show the exact parameter combinations in \cref{app:benchmark_params} and discuss details in \cref{app:implementation_details}. Additionally, we do not optimally use our codebooks: We do not ensure that our codebook resets are optimal, as we do not ensure that we split a codebook entry in an optimal way (\cref{fig:reassignment} shows an optimal split). This can lead to potentially having multiple codewords that are very similar, but are still each with a high gradient. We only reset unused codebook entries instead of resetting very rarely used ones. Also, we decided not to include attention\cite{attention}, as to not dilute that the \textit{architecture itself} is able to produce a global representation. We leave this for future work to keep our baseline approach simple. 

%% file: sections/figures/examples.tex
\begin{figure}[htbp]
    \centering
    \includegraphics[width=1.0\linewidth]{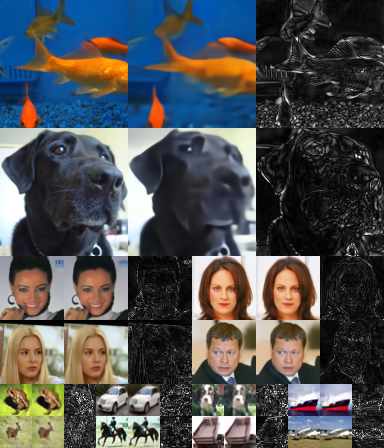}
    \caption{Input, output, and absolute difference between the two, produced by our autoencoder. $256$ tokens/$288$ byte for ImageNet\cite{imagenet} and CelebA\cite{celeba}, $64$ tokens/$72$ bytes for CIFAR-10\cite{cifar}. All with $512$ tokens in the codebook. Note how sharp the features, \eg the horse rider, in CIFAR are.}
    \label{fig:ae_examples}
\end{figure}

%% file: sections/tables/compression.tex
\begin{table}[t!h]
\centering
\begin{tabular}{|c||c|c|c||c|}
\hline
\multicolumn{5}{|c|}{\textbf{MNIST}\cite{mnist}, $8$-by-$8$ tokens versus $64$ tokens (\textbf{ours})} \\
\hline
Metric & VQ-VAE & SVQ-VAE& VQ-WAE & QG-VAE \\
\hline
SSIM $\uparrow$ & 0.98  & 0.99 & 0.99 & \textbf{0.997}\\
PSNR $\uparrow$ & 33.37  & 36.25 & 35.71 & \textbf{38.29}\\
LPIPS $\downarrow$ & 0.02  & 0.01 & 0.01 & \textbf{0.002}\\
FID $\downarrow$ & 4.8 & 3.2 & 2.33 & \textbf{1.13}\\
\hline
\hline
\multicolumn{5}{|c|}{\textbf{CIFAR-10}\cite{cifar}, $8$-by-$8$ tokens versus $64$ tokens (\textbf{ours})} \\
\hline
Metric & VQ-VAE & SVQ-VAE & VQ-WAE & QG-VAE \\
\hline
SSIM $\uparrow$ & 0.70 & 0.80 & 0.80 & \textbf{0.86}\\
PSNR $\uparrow$ & 23.14  & 26.11 & 25.93 & \textbf{26.18}\\
LPIPS $\downarrow$ & 0.35  & 0.23 & 0.23 & \textbf{0.18}\\
FID $\downarrow$ & 77.3 & 55.4 & 54.3 & \textbf{38.9} \\
\hline
\hline
\multicolumn{5}{|c|}{\textbf{SVHN}\cite{svhn}, $8$-by-$8$ tokens versus $64$ tokens (\textbf{ours})} \\
\hline
Metric & VQ-VAE & SVQ-VAE & VQ-WAE & QG-VAE \\
\hline
SSIM $\uparrow$ & 0.88  & 0.96 & 0.96 & \textbf{0.97}\\
PSNR $\uparrow$ & 26.94 &  35.37 & 34.62 & \textbf{36.25}\\
LPIPS $\downarrow$ & 0.17  & \textbf{0.06} & 0.07 & \textbf{0.06}\\
FID $\downarrow$ &  38.5 & 24.8 & 23.4 & \textbf{14.16} \\
\hline
\hline
\multicolumn{5}{|c|}{\textbf{CelebA}\cite{celeba}, $16$-by-$16$ tokens versus $256$ tokens (ours)}\\
\hline
Metric &VQ-VAE & SVQ-VAE & VQ-WAE & QG-VAE \\
\hline
SSIM $\uparrow$ & 0.82  & 0.89 & 0.89 &\textbf{0.94}\\
PSNR $\uparrow$ & 27.48  & 31.05 & 30.60 & \textbf{31.44}\\
LPIPS $\downarrow$  & 0.17 & \textbf{0.06} & 0.07 & 0.09 \\
FID $\downarrow$ & 19.4 & 14.8 & \textbf{12.2} & 16.09 \\
\hline
\end{tabular}
\caption{Comparison of our approach (QG-VAE) to other VQ-VAE-based compression schemes in similar conditions, namely the vanilla VQ-VAE\cite{vqvae}, SVQ-VAE\cite{svqvae}, and VQ-WAE\cite{vqwae}. We build upon the data of \cite{vqwae}. Details given in \cref{app:benchmark_params}. Bold indicates best performance.}\label{tab:compression}
\end{table}

%% file: sections/figures/qggan.tex
\begin{figure}[h!]
    \centering
    \begin{tabular}{cccc}
        Input & QG-VAE (ours) & QGGAN (ours)\\
        \\
        \subfloat{\includegraphics[width=0.14\textwidth]{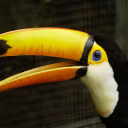}}
        &\subfloat{\includegraphics[width=0.14\textwidth]{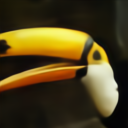}} &\subfloat{\includegraphics[width=0.14\textwidth]{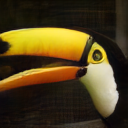}} \\
        \subfloat{\includegraphics[width=0.14\textwidth]{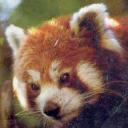}}
        &\subfloat{\includegraphics[width=0.14\textwidth]{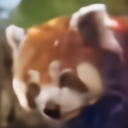}} &\subfloat{\includegraphics[width=0.14\textwidth]{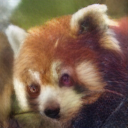}} \\
        \subfloat{\includegraphics[width=0.14\textwidth]{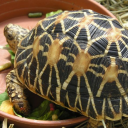}}
        &\subfloat{\includegraphics[width=0.14\textwidth]{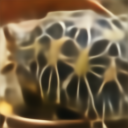}} &\subfloat{\includegraphics[width=0.14\textwidth]{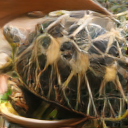}} \\
    \end{tabular}
    \caption{Examples on ImageNet $128$ with sharpening, from left to right: Input, output (QG-VAE), output (QGGAN). Each image has only $256$ tokens ($288$ bytes).}\label{fig:images_qggan}
\end{figure}

%% file: sections/figures/dropout.tex
\begin{figure*}[http]
    \begin{minipage}[t]{0.75\textwidth}
        \centering
        \begin{minipage}[t]{0.19\textwidth}
            \centering
            \includegraphics[width=\textwidth]{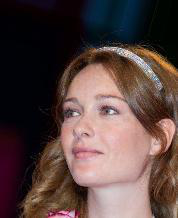}
            \caption*{Input}
        \end{minipage}
        \hfill
        \begin{minipage}[t]{0.19\textwidth}
            \centering
            \includegraphics[width=\textwidth]{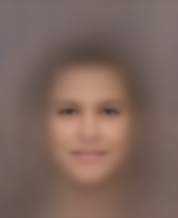}
            \caption*{$0$ tokens}
        \end{minipage}
        \hfill
        \begin{minipage}[t]{0.19\textwidth}
            \centering
            \includegraphics[width=\textwidth]{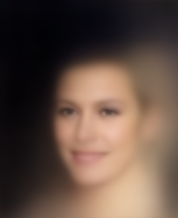}
            \caption*{First $8$ tokens}
        \end{minipage}
        \hfill
        \begin{minipage}[t]{0.19\textwidth}
            \centering
            \includegraphics[width=\textwidth]{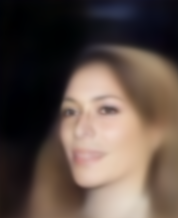}
            \caption*{First $32$ tokens}
        \end{minipage}
        \hfill
        \begin{minipage}[t]{0.19\textwidth}
            \centering
            \includegraphics[width=\textwidth]{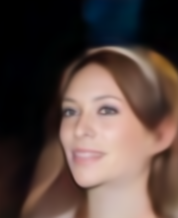}
            \caption*{First $128$ tokens}
        \end{minipage}
        \caption{Learned pseudo frequency decomposition on the CelebA dataset \cite{celeba}. From left: Input, then from left to right, we allow more tokens in the latent code to obtain more detail in the output. All images produced by using the regularisation from \cref{sec:dropout}.}
        \label{fig:dropout}
    \end{minipage}
    \begin{minipage}[t]{0.24\textwidth}
        \centering
        \includegraphics[width=\textwidth]{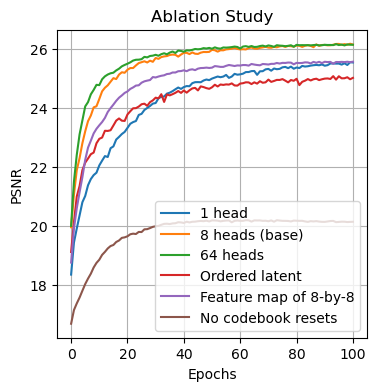}
        \caption{A small ablation study on CIFAR-10\cite{cifar}, showing various settings and the impact of our regularisation procedure to order the latent space.}\label{fig:ablation}
    \end{minipage}
\end{figure*}

%% file: sections/figures/featuremaps.tex
\begin{figure}[h!]
    \centering
    \begin{tabular}{cccccccc}
        \subfloat{\includegraphics[width=0.075\textwidth]{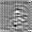}} &
        \subfloat{\includegraphics[width=0.075\textwidth]{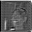}} &
        \subfloat{\includegraphics[width=0.075\textwidth]{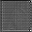}} &
        \subfloat{\includegraphics[width=0.075\textwidth]{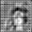}} &
        \subfloat{\includegraphics[width=0.075\textwidth]{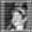}} \\
        \subfloat{\includegraphics[width=0.075\textwidth]{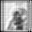}} &
        \subfloat{\includegraphics[width=0.075\textwidth]{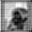}} &
        \subfloat{\includegraphics[width=0.075\textwidth]{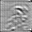}} &
        \subfloat{\includegraphics[width=0.075\textwidth]{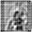}} &
        \subfloat{\includegraphics[width=0.075\textwidth]{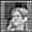}} \\
        \subfloat{\includegraphics[width=0.075\textwidth]{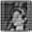}} &
        \subfloat{\includegraphics[width=0.075\textwidth]{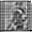}} &
        \subfloat{\includegraphics[width=0.075\textwidth]{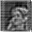}} &
        \subfloat{\includegraphics[width=0.075\textwidth]{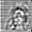}} &
        \subfloat{\includegraphics[width=0.075\textwidth]{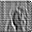}} 
    \end{tabular}
    \caption{Learned feature maps of our approach: We normalised the individual $C$ many $\frac{W}{2^f}$-by-$\frac{H}{2^f}$ feature maps at the layer before they would be compressed into pseudo frequency tokens by the affine transformation. We then visualised $16$ randomly chosen ones here, all from the same image as \cref{fig:dropout}, but not ordered, \ie without \cref{sec:dropout}.}\label{fig:feature_maps}
\end{figure}

%% file: sections/5_conclusion.tex
\section{Conclusion}
We propose a holistic approach to autoencoding, capturing global features in quantised tokens. Opposing to traditional decompositions, we process these pseudo frequency tokens together. We believe that this holistic take on autoencoding has the potential to spawn a whole new branch of autoencoders, as we demonstrate success in multiple benchmarks with a rather simple formulation. We do so without any powerful mathematical regularisations like Wasserstein loss or recent architectural improvements like attention. We further demonstrate that our approach improves generation significantly, and that it is compatible to autoencoder improvements like additional sharpening.

%% file: sections/6_appendix.tex
\section{Benchmark Parameters}\label{app:benchmark_params}
While modern optimisers such as schedule free Adam\cite{sfadam} worked better in our early experiments, for comparability, we train all our examples with AdamW\cite{adamw}, using a learning rate of $0.0002$ with weight decay of $0.01$. While the exact number of parameters of the architecture in \cite{vqwae} is not specified, we argue that our architecture is vastly different and hence difficult to compare. However, exemplary for CIFAR, a classic VQ-VAE with the same number of channels performs worse, with similar numbers for the other approaches. Our approach works similarly good when using significantly less parameters when using a higher downscaling factor, \ie the encoder U-Net producing a $[B \times C \times \frac{W}{8} \times \frac{H}{8}]$ as we demonstrate for cases in ImageNet and CelebA in \cref{fig:celeb_big} and \cref{fig:extra_images_imagenet}.\\
We omit the perplexity score in our benchmarks, as we do not, as in VQ-VAE\cite{vqvae}, have the same tokens in different locations, but tokens that can mean very different things in different pseudo frequency bands.\\ 
We specify the exact settings in \cref{tab:app_settings}. Note that for CelebA, for comparability, we follow \cite{vqwae} in using a $64$-by-$64$ version of CelebA.
\begin{table*}[h!]
\centering
\begin{tabular}{|c|c|c|c|c|c|c|c|c|}
\hline
Used in & Dataset & Codewords & Feature Map / $f$ & Heads & Blocks & Channels & Reset time & Epochs \\
\hline\hline
Default & MNIST\cite{cifar} & $64$ & $32$-by-$32 / f=0$ & $8$ & $2$ & $256$ & $100$ & $100$ \\
\hline
Default & CIFAR-10\cite{cifar} & $64$ & $32$-by-$32 / f=0$ & $8$ & $1$ & $128$ & $100$ & $100$ \\
\hline
Default & SVHN\cite{svhn} & $64$ & $32$-by-$32 / f=0$ & $8$ & $1$ & $128$ & $100$ & $100$ \\
\hline
Default & CelebA(64)\cite{celeba} & $256$ & $16$-by-$16$ / $f=2$ & $64$ & $2$ & $256$ & $1000$ & $70$ \\
\hline
Default & CelebA(256*)\cite{celeba} & $256$ & $32$-by-$32$ / $f=3$ & $64$ & $2$ & $128$ & $1350$ & $55$ \\
\hline
\cref{fig:app_imagenet64} & ImageNet$64$\cite{imagenet} & 256/1024 & $16$-by-$16$ / $f=2$ & $64$ & $2$ & $128$ & $1000$ & $100$ \\
\hline
Default & ImageNet$128$\cite{imagenet} & $256$ & $32$-by-$32$ / $f=2$ & $64$ & $2$ & $128$ & $1000$ & $15$ \\
\hline
QGGAN & ImageNet$128$\cite{imagenet} & $256$ & $16$-by-$16$ / $f=3$ & $128$ & $2$ & $128$ & $1000$ & $15$ \\
\hline
\end{tabular}
\caption{The different training parameters used for our experiments, with 'Default' meaning these settings were used throughout the paper if not stated otherwise. 'Codewords' describes the number of codewords that encode one input, 'Feature Map' the resulting width/height after the U-Net, 'Heads' shows the number of different affine transformations to turn a feature map into the space in which we quantise, 'Blocks' and 'Channels' describe the number of residual blocks and maximum number of channels for our U-Net, and 'Reset time' describes after how many batches we reset our unused codebook entries.\\
*We trained on the original CelebA images embedded in $256$ pixel black boxes to avoid adapting our U-Net. As we have no local tokens, adding a static black frame around our image does not harm quality.}
\label{tab:app_settings}
\end{table*}

\section{Implementation Details}\label{app:implementation_details}
As the number of codewords grow, \eg from a very large input image in full HD resolution, the number of codebooks grows with it. While this has not been an issue for any of our benchmarks or even our large test images on ImageNet\cite{imagenet} at $256$ pixels, we recommend using either fewer dimensions for the quantisation dimension or using partially shared codebooks between \eg $16$ frequencies at a time, similar to the heads of the affine transformation that project an $[B \times C \times W \times H]$-sized feature map to the quantisation dimension. We observed this to work well in early experiments, but omitted this due not not processing any extra large images and slightly weaker codebook resets.

\section{Additional Feature Maps}
\begin{figure}[h!]
    \centering
    \begin{tabular}{cccc}
        \subfloat{\includegraphics[width=0.1\textwidth]{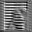}} &
        \subfloat{\includegraphics[width=0.1\textwidth]{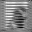}} &
        \subfloat{\includegraphics[width=0.1\textwidth]{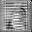}} &
        \subfloat{\includegraphics[width=0.1\textwidth]{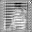}} \\
        \subfloat{\includegraphics[width=0.1\textwidth]{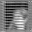}} &
        \subfloat{\includegraphics[width=0.1\textwidth]{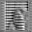}} &
        \subfloat{\includegraphics[width=0.1\textwidth]{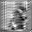}} &
        \subfloat{\includegraphics[width=0.1\textwidth]{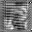}} \\
        \subfloat{\includegraphics[width=0.1\textwidth]{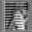}} &
        \subfloat{\includegraphics[width=0.1\textwidth]{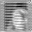}} &
        \subfloat{\includegraphics[width=0.1\textwidth]{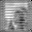}} &
        \subfloat{\includegraphics[width=0.1\textwidth]{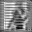}} \\
        \subfloat{\includegraphics[width=0.1\textwidth]{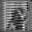}} &
        \subfloat{\includegraphics[width=0.1\textwidth]{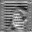}} &
        \subfloat{\includegraphics[width=0.1\textwidth]{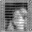}} &
        \subfloat{\includegraphics[width=0.1\textwidth]{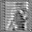}} \\
    \end{tabular}
    \caption{Learned feature maps of our approach when ordering the latent space as discussed in \cref{sec:dropout}: We normalised the individual $C$ many $\frac{W}{2^f}$-by-$\frac{H}{2^f}$ feature maps from the step before they would be compressed into pseudo frequency tokens by the affine transformation, then visualised the first $16$ ones, all from the same image as \cref{fig:dropout}.}\label{fig:features_ordered}
\end{figure}
We provide the ordered feature maps of \cref{fig:dropout}, similarly to \cref{fig:feature_maps}, in \cref{fig:features_ordered}. Note how these feature maps are visibly more orthogonal to each other.

\section{Training Progress}\label{app:training}
\begin{figure*}[h!]
    \centering
    \begin{tabular}{cccc}
        VQ-VAE, Input, QG-VAE & VQ-VAE, Input, QG-VAE & VQ-VAE, Input, QG-VAE & VQ-VAE, Input, QG-VAE\\
        \subfloat{\includegraphics[width=0.23\textwidth]{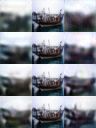}} &
        \subfloat{\includegraphics[width=0.23\textwidth]{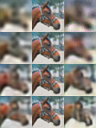}} &
        \subfloat{\includegraphics[width=0.23\textwidth]{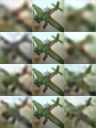}} &
        \subfloat{\includegraphics[width=0.23\textwidth]{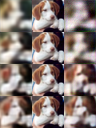}}
    \end{tabular}
    \caption{Slightly different formation of features during training. The figure is meant to show the different ways how the two approaches learn, not to compare the image quality. Therefore, all images in a row are taken during roughly the same PSNR, with increasing PSNR from top to bottom.}\label{fig:app_trainprocess}
\end{figure*}
Differences during the way an image forms during training are observable, we hence output a number of processes in \cref{fig:app_trainprocess}. We can observe that our approach learns to form the image slightly differently.

\section{Alternative Formulations}\label{app:alternatives}
While our approach of first reducing an image size, then translating between feature and channel dimension to turn local into global features, works well, we also considered some alternative ways to obtain global tokens:
\paragraph{Fully Convolutional Architecture} For an input image, we can compress it down by applying convolution and pooling layers until we reach a single feature with $C$ many channels to represent an image, then quantise and decode the result. However, as the quantisation usually needs about $64$ dimensions to work properly (see \cite{vqvae}), this would require $C * 64$ many channels at feature size $1$-by-$1$. For \eg $1024$ tokens (which we need for proper ImageNet reconstruction quality in $256$ pixels), this would result in $65,536$ channels, with the previous layer thus requiring at least to be $[B \times 16,384 \times 2 \times 2]$. In turn, the operation to reduce from $16,384$ to $65,536$ channels would require over a billion parameters, making this approach not feasible for any larger image.
\paragraph{Fully Attention Based} Again, for larger images like ImageNet with $256$ pixels, this would require computation of $n^2 = 65,536$ many attention scores. While this is certainly possible \eg on an H$100$ GPU, this is often not feasible for labs with in-expensive hardware.
\paragraph{Fully Linear Architecture} Staying with the ImageNet example, even when applying a few layers of convolution and pooling first, when combining \eg $16$-by-$16$ feature dimensions with $256$ channels into $64 \cdot 1024$ (to have enough space for quantisation and to produce $1024$ tokens), this again would require over $4$ billion parameters. Again, this is not feasible in the sense of both hardware requirements and generalisation.
\section{Smaller Experiments}\label{app:smaller}
\begin{figure}
    \centering
    \includegraphics[width=0.25\textwidth]{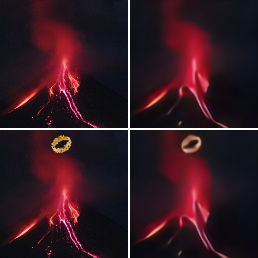}
    \caption{Even the smallest change can change the course of the representation: More than $10$ percent of the tokens are changed by adding the ring to the volcano. Left: Input, Right: Output.}\label{fig:ring}
\end{figure}

We can apply local changes, like adding a ring to a volcano, to provoke large changes in token space, as can be seen in \cref{fig:ring} (left shows input, right shows output). 

\section{Additional Results, Qualitative Analysis, Comparison Against VQ-VAE}\label{app:extra_results}
\begin{figure*}
    \centering
    \includegraphics[width=1.0\textwidth]{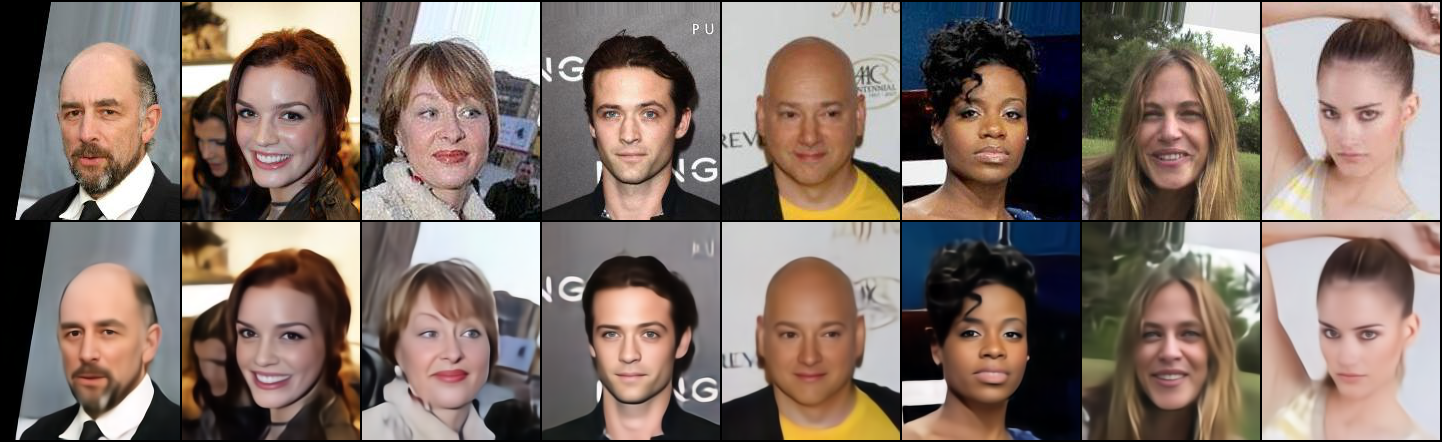}
    \caption{Additional results of CelebA, but in full size, with inputs (top) and outputs (bottom). We use twice as many tokens here as in the $64$-by-$64$ benchmarks in \cref{tab:compression} ($512$). We report a PSNR of $\sim28.2$.}\label{fig:celeb_big}
\end{figure*}

We provide additional examples of our autoencoder quality, in particular on larger datasets and higher resolutions. In \cref{fig:celeb_big}, we show results of CelebA\cite{celeba} in original size (opposing to the $64$-by-$64$ results used for the benchmarks here and in originally in \cite{vqwae}). To keep our U-Net simple, we embed the $178$-by-$218$ in black $256^2$ images. Conveniently, due to global instead of local tokens, our approach does not suffer from unused boundary regions that we throw away at the end.\\
\paragraph{ImageNet} While we did not extensive train and fine-tune our approach on ImageNet\cite{imagenet} due to a lack of benchmarks and resources (\eg no comparison to VQ-WAE\cite{vqwae} possible due to undisclosed code), we do provide some results on (unoptimised) settings to demonstrate that we can also perform well on larger datasets and with larger numbers of tokens. We train on ImageNet in $64$-by-$64$ pixels with $256$ and $1024$ tokens. We also train our approach on $128$-by-$128$ pixels with $256$ tokens, shown in \cref{fig:extra_images_imagenet}, comparing against an already improved VQ-VAE with codebook regularisation that avoids unused codebook entries. In result, our approach is better at producing sharper features, even if they are at times slightly misplaced. We also give additional examples produced by QGGAN, \ie a QG-VAE with additional VQGAN\cite{vqgan}-like sharpening.

\section{Generative Results}\label{app:generation}
\begin{figure}
    \centering
    \includegraphics[width=0.48\textwidth]{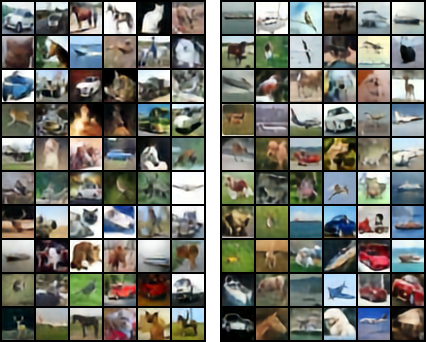}
    \caption{Results of applying a generative transformers for unconditional generation on our learned latent space. Left: VQ-VAE for reference, right: QG-VAE (ours). The transformer applied to tokens produced by ours has fewer 'empty' backgrounds, which we attribute to our holistic and global approach to image representation: Our tokens prohibit the 'easy' solution of just placing multiple uniform coloured tokens in sequence.}\label{fig:app_generative_results}
\end{figure}

We provide additional randomly selected result of random autoregressive generation of our QGVAE in \cref{fig:app_generative_results}.

\input{sections/figures/examples_imagenet128}
\begin{figure*}
    \centering
    \includegraphics[width=1.0\textwidth]{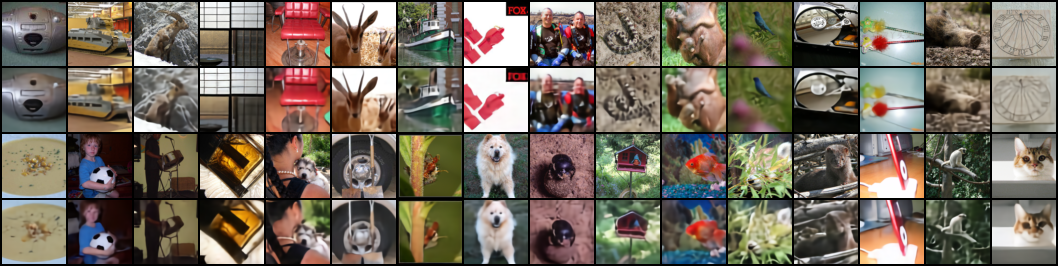}
    \includegraphics[width=1.0\textwidth]{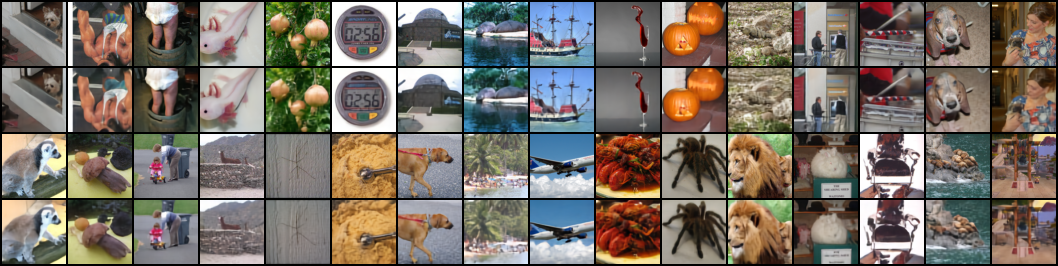}
    \caption{Additional results on ImageNet64, with inputs always above the outputs, with $256$ tokens ($288$ bytes per image, upper half of the figure). We report a PSNR of $\sim 28.4$ after $4$ epochs. For $1024$ tokens, we obtain a PSNR of $\sim 31.4$ after 4 epochs (lower half of the figure).}\label{fig:app_imagenet64}
\end{figure*}

%% file: sections/figures/examples_imagenet128.tex
\begin{figure*}[h!]
    \centering
    \begin{tabular}{cccc}
        Input & VQ-VAE & QG-VAE (ours) & QGGAN (ours)\\
        \subfloat{\includegraphics[width=0.2\textwidth]{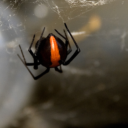}}
        &\subfloat{\includegraphics[width=0.2\textwidth]{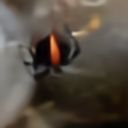}} &\subfloat{\includegraphics[width=0.2\textwidth]{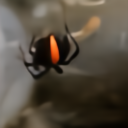}} &\subfloat{\includegraphics[width=0.2\textwidth]{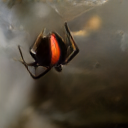}}\\
        \subfloat{\includegraphics[width=0.2\textwidth]{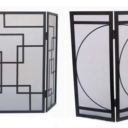}} &\subfloat{\includegraphics[width=0.2\textwidth]{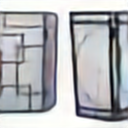}} &\subfloat{\includegraphics[width=0.2\textwidth]{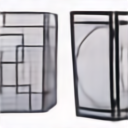}} &\subfloat{\includegraphics[width=0.2\textwidth]{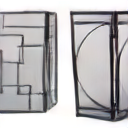}}\\
        \subfloat{\includegraphics[width=0.2\textwidth]{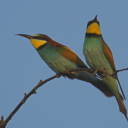}}
        &\subfloat{\includegraphics[width=0.2\textwidth]{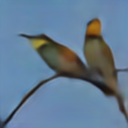}} &\subfloat{\includegraphics[width=0.2\textwidth]{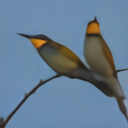}} &\subfloat{\includegraphics[width=0.2\textwidth]{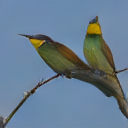}} \\
        \subfloat{\includegraphics[width=0.2\textwidth]{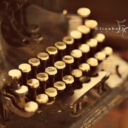}} &\subfloat{\includegraphics[width=0.2\textwidth]{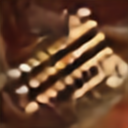}} &\subfloat{\includegraphics[width=0.2\textwidth]{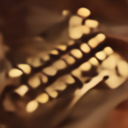}} &\subfloat{\includegraphics[width=0.2\textwidth]{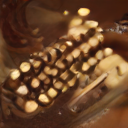}}\\
        \subfloat{\includegraphics[width=0.2\textwidth]{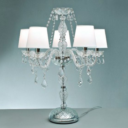}}
        &\subfloat{\includegraphics[width=0.2\textwidth]{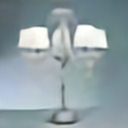}} 
        &\subfloat{\includegraphics[width=0.2\textwidth]{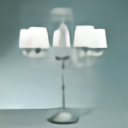}} 
        & \subfloat{\includegraphics[width=0.2\textwidth]{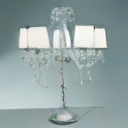}} \\
        \subfloat{\includegraphics[width=0.2\textwidth]{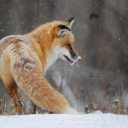}} 
        &\subfloat{\includegraphics[width=0.2\textwidth]{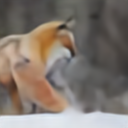}} 
        &\subfloat{\includegraphics[width=0.2\textwidth]{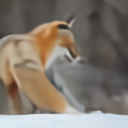}} 
        &\subfloat{\includegraphics[width=0.2\textwidth]{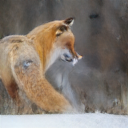}}
    \end{tabular}
\end{figure*}
\begin{figure*}[h!]
    \centering
    \begin{tabular}{cccc}
        Input & VQ-VAE & QG-VAE (ours) & QGGAN (ours)\\
        \\
        \subfloat{\includegraphics[width=0.2\textwidth]{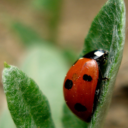}}
        &\subfloat{\includegraphics[width=0.2\textwidth]{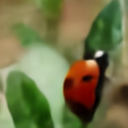}} &\subfloat{\includegraphics[width=0.2\textwidth]{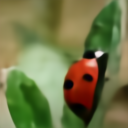}} &\subfloat{\includegraphics[width=0.2\textwidth]{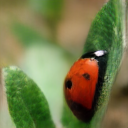}} \\
        \subfloat{\includegraphics[width=0.2\textwidth]{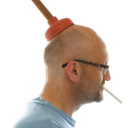}} &\subfloat{\includegraphics[width=0.2\textwidth]{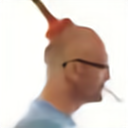}} &\subfloat{\includegraphics[width=0.2\textwidth]{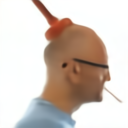}} &\subfloat{\includegraphics[width=0.2\textwidth]{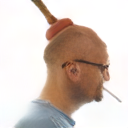}}\\
        \subfloat{\includegraphics[width=0.2\textwidth]{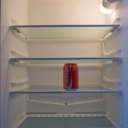}}
        &\subfloat{\includegraphics[width=0.2\textwidth]{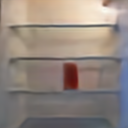}} &\subfloat{\includegraphics[width=0.2\textwidth]{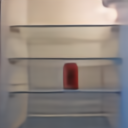}}  &\subfloat{\includegraphics[width=0.2\textwidth]{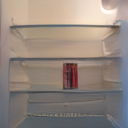}}\\ 
        \subfloat{\includegraphics[width=0.2\textwidth]{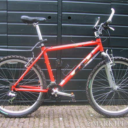}} &\subfloat{\includegraphics[width=0.2\textwidth]{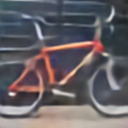}} &\subfloat{\includegraphics[width=0.2\textwidth]{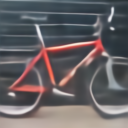}} &\subfloat{\includegraphics[width=0.2\textwidth]{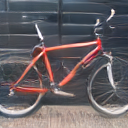}}\\
        \subfloat{\includegraphics[width=0.2\textwidth]{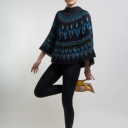}}
        &\subfloat{\includegraphics[width=0.2\textwidth]{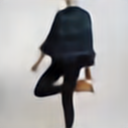}} &\subfloat{\includegraphics[width=0.2\textwidth]{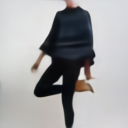}}  &\subfloat{\includegraphics[width=0.2\textwidth]{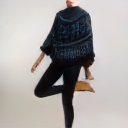}}\\ 
        \subfloat{\includegraphics[width=0.2\textwidth]{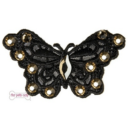}} &\subfloat{\includegraphics[width=0.2\textwidth]{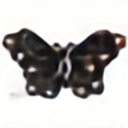}} &\subfloat{\includegraphics[width=0.2\textwidth]{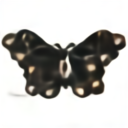}} &\subfloat{\includegraphics[width=0.2\textwidth]{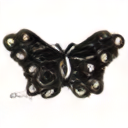}}
    \end{tabular}
    \caption{Examples on ImageNet $128$, always input, output (VQ-VAE\cite{vqvae}), output (QG-VAE), output (QGGAN). Note how small details, like the spider legs or global structures like the fence behind the bike, are better represented for our approach instead of just being blurred. Each image has only $256$ tokens ($288$ bytes).}\label{fig:extra_images_imagenet}
\end{figure*}

%% file: Main.bbl
\begin{thebibliography}{10}\itemsep=-1pt

\bibitem{attention}
Dzmitry Bahdanau, Kyunghyun Cho, and Yoshua Bengio.
\newblock Neural machine translation by jointly learning to align and
  translate.
\newblock {\em arXiv preprint arXiv:1409.0473}, 2014.

\bibitem{vq_optimal_transport}
Xiaoyu Bie, Dexiong Chen, Xiaodong Cun, and Xi SHEN.
\newblock Learning discrete representation with optimal transport quantized
  autoencoders, 2023.

\bibitem{laplacian}
Peter~J Burt and Edward~H Adelson.
\newblock The laplacian pyramid as a compact image code.
\newblock In {\em Readings in computer vision}, pages 671--679. Elsevier, 1987.

\bibitem{sfadam}
Aaron Defazio, Xingyu, Yang, Harsh Mehta, Konstantin Mishchenko, Ahmed Khaled,
  and Ashok Cutkosky.
\newblock The road less scheduled, 2024.

\bibitem{imagenet}
Jia Deng, Wei Dong, Richard Socher, Li-Jia Li, Kai Li, and Li Fei-Fei.
\newblock Imagenet: A large-scale hierarchical image database.
\newblock In {\em 2009 IEEE conference on computer vision and pattern
  recognition}, pages 248--255. Ieee, 2009.

\bibitem{vit}
Alexey Dosovitskiy, Lucas Beyer, Alexander Kolesnikov, Dirk Weissenborn,
  Xiaohua Zhai, Thomas Unterthiner, Mostafa Dehghani, Matthias Minderer, Georg
  Heigold, Sylvain Gelly, Jakob Uszkoreit, and Neil Houlsby.
\newblock An image is worth 16x16 words: Transformers for image recognition at
  scale.
\newblock {\em CoRR}, abs/2010.11929, 2020.

\bibitem{vqgan}
Patrick Esser, Robin Rombach, and Bj{\"{o}}rn Ommer.
\newblock Taming transformers for high-resolution image synthesis.
\newblock {\em CoRR}, abs/2012.09841, 2020.

\bibitem{gan}
Ian Goodfellow, Jean Pouget-Abadie, Mehdi Mirza, Bing Xu, David Warde-Farley,
  Sherjil Ozair, Aaron Courville, and Yoshua Bengio.
\newblock Generative adversarial nets.
\newblock In Z. Ghahramani, M. Welling, C. Cortes, N. Lawrence, and K.Q.
  Weinberger, editors, {\em Advances in Neural Information Processing Systems},
  volume~27. Curran Associates, Inc., 2014.

\bibitem{vec_gen}
Shuyang Gu, Dong Chen, Jianmin Bao, Fang Wen, Bo Zhang, Dongdong Chen, Lu Yuan,
  and Baining Guo.
\newblock Vector quantized diffusion model for text-to-image synthesis.
\newblock In {\em Proceedings of the IEEE/CVF conference on computer vision and
  pattern recognition}, pages 10696--10706, 2022.

\bibitem{heeger}
David~J Heeger and James~R Bergen.
\newblock Pyramid-based texture analysis/synthesis.
\newblock In {\em Proceedings of the 22nd annual conference on Computer
  graphics and interactive techniques}, pages 229--238, 1995.

\bibitem{hinton2006reducing}
Geoffrey~E Hinton and Ruslan~R Salakhutdinov.
\newblock Reducing the dimensionality of data with neural networks.
\newblock {\em science}, 313(5786):504--507, 2006.

\bibitem{hier_vqvae}
Mengqi Huang, Zhendong Mao, Zhuowei Chen, and Yongdong Zhang.
\newblock Towards accurate image coding: Improved autoregressive image
  generation with dynamic vector quantization, 2023.

\bibitem{mingpt}
Andrew Karpathy.
\newblock mingpt, 2023.

\bibitem{vae}
Diederik~P Kingma and Max Welling.
\newblock Auto-encoding variational bayes, 2022.

\bibitem{cifar}
Alex Krizhevsky, Geoffrey Hinton, et~al.
\newblock Learning multiple layers of features from tiny images.
\newblock 2009.

\bibitem{lancucki2020robust}
Adrian {\L}a{\'n}cucki, Jan Chorowski, Guillaume Sanchez, Ricard Marxer, Nanxin
  Chen, Hans~JGA Dolfing, Sameer Khurana, Tanel Alum{\"a}e, and Antoine
  Laurent.
\newblock Robust training of vector quantized bottleneck models.
\newblock In {\em 2020 International Joint Conference on Neural Networks
  (IJCNN)}, pages 1--7. IEEE, 2020.

\bibitem{mnist}
Yann LeCun.
\newblock The mnist database of handwritten digits.
\newblock {\em http://yann. lecun. com/exdb/mnist/}, 1998.

\bibitem{celeba}
Ziwei Liu, Ping Luo, Xiaogang Wang, and Xiaoou Tang.
\newblock Deep learning face attributes in the wild.
\newblock In {\em Proceedings of International Conference on Computer Vision
  (ICCV)}, December 2015.

\bibitem{adamw}
Ilya Loshchilov and Frank Hutter.
\newblock Decoupled weight decay regularization, 2019.

\bibitem{simple_vq}
Fabian Mentzer, David Minnen, Eirikur Agustsson, and Michael Tschannen.
\newblock Finite scalar quantization: Vq-vae made simple.
\newblock {\em arXiv preprint arXiv:2309.15505}, 2023.

\bibitem{svhn}
Yuval Netzer, Tao Wang, Adam Coates, Alessandro Bissacco, Baolin Wu, Andrew~Y
  Ng, et~al.
\newblock Reading digits in natural images with unsupervised feature learning.
\newblock In {\em NIPS workshop on deep learning and unsupervised feature
  learning}, volume 2011, page~4. Granada, 2011.

\bibitem{dalle}
Aditya Ramesh, Mikhail Pavlov, Gabriel Goh, Scott Gray, Chelsea Voss, Alec
  Radford, Mark Chen, and Ilya Sutskever.
\newblock Zero-shot text-to-image generation.
\newblock {\em CoRR}, abs/2102.12092, 2021.

\bibitem{vqvae2}
Ali Razavi, Aaron van~den Oord, and Oriol Vinyals.
\newblock Generating diverse high-fidelity images with vq-vae-2, 2019.

\bibitem{unet}
Olaf Ronneberger, Philipp Fischer, and Thomas Brox.
\newblock U-net: Convolutional networks for biomedical image segmentation.
\newblock In {\em Medical image computing and computer-assisted
  intervention--MICCAI 2015: 18th international conference, Munich, Germany,
  October 5-9, 2015, proceedings, part III 18}, pages 234--241. Springer, 2015.

\bibitem{first_ae}
David~E Rumelhart, Geoffrey~E Hinton, and Ronald~J Williams.
\newblock Learning internal representations by error propagation, parallel
  distributed processing, explorations in the microstructure of cognition, ed.
  de rumelhart and j. mcclelland. vol. 1. 1986.
\newblock {\em Biometrika}, 71(599-607):6, 1986.

\bibitem{imagen}
Chitwan Saharia, William Chan, Saurabh Saxena, Lala Li, Jay Whang, Emily~L
  Denton, Kamyar Ghasemipour, Raphael Gontijo~Lopes, Burcu Karagol~Ayan, Tim
  Salimans, et~al.
\newblock Photorealistic text-to-image diffusion models with deep language
  understanding.
\newblock {\em Advances in neural information processing systems},
  35:36479--36494, 2022.

\bibitem{sgan}
Jürgen Schmidhuber.
\newblock Learning factorial codes by predictability minimization.
\newblock {\em Neural Computation}, 4(6):863--879, 1992.

\bibitem{arch_imp_1}
Dario Serez, Marco Cristani, Vittorio Murino, Alessio Del~Bue, and Pietro
  Morerio.
\newblock Enhancing hierarchical vector quantized autoencoders for image
  synthesis through multiple decoders.
\newblock In {\em International Conference on Image Analysis and Processing},
  pages 393--405. Springer, 2023.

\bibitem{svqvae}
Marek Strong, Jonas Rohnke, Antonio Bonafonte, Mateusz {\L}ajszczak, and Trevor
  Wood.
\newblock Discrete acoustic space for an efficient sampling in neural
  text-to-speech.
\newblock {\em arXiv preprint arXiv:2110.12539}, 2021.

\bibitem{eigenfaces}
M.A. Turk and A.P. Pentland.
\newblock Face recognition using eigenfaces.
\newblock In {\em Proceedings. 1991 IEEE Computer Society Conference on
  Computer Vision and Pattern Recognition}, pages 586--591, 1991.

\bibitem{vqvae}
Aaron Van Den~Oord, Oriol Vinyals, et~al.
\newblock Neural discrete representation learning.
\newblock {\em Advances in neural information processing systems}, 30, 2017.

\bibitem{transformer}
Ashish Vaswani, Noam Shazeer, Niki Parmar, Jakob Uszkoreit, Llion Jones,
  Aidan~N. Gomez, Lukasz Kaiser, and Illia Polosukhin.
\newblock Attention is all you need, 2023.

\bibitem{vqwae}
Tung-Long Vuong, Trung Le, He Zhao, Chuanxia Zheng, Mehrtash Harandi, Jianfei
  Cai, and Dinh Phung.
\newblock Vector quantized wasserstein auto-encoder, 2023.

\bibitem{jpg}
G.K. Wallace.
\newblock The jpeg still picture compression standard.
\newblock {\em IEEE Transactions on Consumer Electronics}, 38(1):xviii--xxxiv,
  1992.

\bibitem{comp_cosine}
Andrew~B Watson et~al.
\newblock Image compression using the discrete cosine transform.
\newblock {\em Mathematica journal}, 4(1):81, 1994.

\bibitem{wu2020vector}
Hanwei Wu and Markus Flierl.
\newblock Vector quantization-based regularization for autoencoders.
\newblock In {\em Proceedings of the AAAI Conference on Artificial
  Intelligence}, volume~34, pages 6380--6387, 2020.

\bibitem{vq_video}
Wilson Yan, Yunzhi Zhang, Pieter Abbeel, and Aravind Srinivas.
\newblock Videogpt: Video generation using vq-vae and transformers.
\newblock {\em arXiv preprint arXiv:2104.10157}, 2021.

\bibitem{vqgan2}
Jiahui Yu, Xin Li, Jing~Yu Koh, Han Zhang, Ruoming Pang, James Qin, Alexander
  Ku, Yuanzhong Xu, Jason Baldridge, and Yonghui Wu.
\newblock Vector-quantized image modeling with improved {VQGAN}.
\newblock {\em CoRR}, abs/2110.04627, 2021.

\bibitem{titok}
Qihang Yu, Mark Weber, Xueqing Deng, Xiaohui Shen, Daniel Cremers, and
  Liang-Chieh Chen.
\newblock An image is worth 32 tokens for reconstruction and generation.
\newblock {\em arXiv preprint arXiv:2406.07550}, 2024.

\bibitem{cvqvae}
Chuanxia Zheng and Andrea Vedaldi.
\newblock Online clustered codebook.
\newblock In {\em Proceedings of the IEEE/CVF International Conference on
  Computer Vision}, pages 22798--22807, 2023.

\bibitem{vqgan_better}
Lei Zhu, Fangyun Wei, Yanye Lu, and Dong Chen.
\newblock Scaling the codebook size of vqgan to 100,000 with a utilization rate
  of 99\%.
\newblock {\em arXiv preprint arXiv:2406.11837}, 2024.

\end{thebibliography}
